\documentclass[letterpaper, 10 pt, conference]{ieeeconf}  %

\IEEEoverridecommandlockouts                              %
        
\usepackage{cite}
\usepackage{amsmath,amssymb,amsfonts}

\usepackage{arydshln}
\usepackage{graphicx}
\usepackage{textcomp}
\usepackage{xcolor}
\usepackage{gensymb}
\usepackage{multirow}
\usepackage{algorithm}
\usepackage{algpseudocode}
\usepackage{amsmath}
\usepackage{pifont}
\DeclareMathOperator*{\argmax}{arg\,max}
\def\BibTeX{{\rm B\kern-.05em{\sc i\kern-.025em b}\kern-.08em
    T\kern-.1667em\lower.7ex\hbox{E}\kern-.125emX}}
\usepackage{mathtools}

\usepackage{siunitx}
\usepackage{url}
\usepackage{overpic}
\usepackage{pgf}
\usepackage{microtype}
\usepackage{booktabs}
\usepackage{todonotes}
\usepackage{color, colortbl}  %
\usepackage{hyperref}

\usepackage{subcaption}
\def\eqref#1{Eq.~(\ref{#1})}

\newcolumntype{C}[1]{>{\centering\let\newline\\\arraybackslash\hspace{0pt}}m{#1}}
\newcolumntype{R}[1]{>{\raggedleft\let\newline\\\arraybackslash\hspace{0pt}}m{#1}}

\title{\LARGE \bf OASIS-Map: Object-Level Change Detection in Multi-Session Mapping using Semantic Correspondence Matching}

\def\WithAuthor{1}
\if\WithAuthor1

\author{
Haedam Oh$^{1}$, Yifu Tao$^{1}$, Nived Chebrolu$^{1,2}$, and Maurice Fallon$^{1}$%
\thanks{$^{1}$Oxford Robotics Institute, Department of Engineering Science, University of Oxford, UK.
Email: \{haedam,yifu,nived,mfallon\}@robots.ox.ac.uk}%
\thanks{$^{2}$Department of Computer Science and Engineering, Indian Institute of Technology Bombay, India.
Email: nived@cse.iitb.ac.in}
}

\else

\author{Anonymous Submission}

\fi

\begin{document}
\bstctlcite{BSTcontrol}

\IEEEaftertitletext{%
\begin{center}
    \centering
	\includegraphics[width=0.99\textwidth]{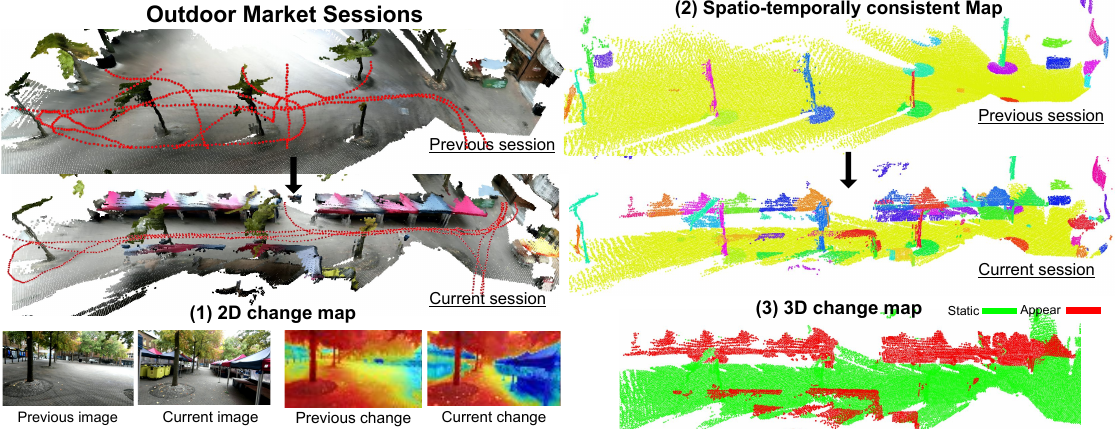}
    \captionof{figure}{Change detection in an outdoor market undergoing a drastic scene transition from an empty square to a fully populated market. Our system
produces three outputs: \textbf{(1)} 2D change maps from dense semantic correspondences; \textbf{(2)} spatio-temporally
consistent maps with consistent object segmentation and association across sessions; and
\textbf{(3)} a 3D object-level change map that tracks the persistency of each
object, e.g.\ newly appeared market tents and tables in red and persistent
structures such as ground and trees in green.  }
    \label{fig:header}
\end{center}
}
\maketitle

\begin{abstract}
Map representations which are consistent across repeated visits to a real-world semi-static environment are very useful for long-term robotic inspection. In such settings, the scene may evolve while the robot is absent, with objects appearing, disappearing, moving, or being replaced, quickly making a static map outdated.
Existing change-detection methods reason through geometry, category-level semantics, or object persistence. However, achieving reliable object association across revisits remains a key challenge, especially under partial views, occlusion, and imperfect segmentation.
In this work, we propose OASIS-Map, a multi-session mapping system that maintains a spatio-temporally consistent object-level map by establishing dense patch-level semantic correspondences between temporal observations. These correspondences detect where the scene has changed and incrementally associate objects across revisits as the robot re-observes the environment.
We demonstrate OASIS-Map on three challenging real-world scenarios: object rearrangements in 3RScan, visually similar car replacements in a car park, and large-scale scene changes in an outdoor market. We achieve 0.783 F1 on change detection in a car replacement scenario in a car park and 0.667 F1 on moved object association in 3RScan.
\url{https://dynamic.robots.ox.ac.uk/projects/oasis-map/}
\end{abstract}

\section{INTRODUCTION}

Inspection robots operating over long periods of time need accurate maps that support localization, asset tracking, and safe planning.
In real-world semi-static environments, the scene continues to evolve while the robot is absent, with objects appearing, disappearing, and moving between visits. In environments such as warehouses, car parks, construction sites or indoor offices, different types of changes occur over different time scales - such as object rearrangement, modification, removal, or replacement.
For reliable long-term autonomy in such environments a robot needs to be able to detect changes, associate them across revisits, and update its map retaining consistency over multiple sessions. The resulting map should be spatio-temporally consistent --- representing not only the current scene, but also how it has evolved over time.

In many semi-static scenes, change is most naturally expressed at the object level: a chair is moved, a car is replaced, or a box disappears from the shelf.
Geometric mapping approaches, such as occupancy grids~\cite{hornung13octomap} and distance fields~\cite{Oleynikova2017Voxblox}, detect change by comparing point-, voxel-, or surface-level differences across sessions~\cite{Fehr2017TSDF,kim2022ltmapper,gil2025elite,Rowell2024LiSTA}.
While effective at identifying where geometry has appeared or disappeared, these methods do not explain \textit{what} has changed, and become ambiguous when objects move slightly or different entities occupy similar locations.
Semantic labeling makes it easier to interpret such changes~\cite{Schmid2022PanopticMultiTSDF}, but category-level semantics still cannot distinguish different instances of the same class, such as two similar cars or boxes. The meaningful unit of change for long-term inspection is therefore not only a semantic category, but also the individual object instance.
This motivates moving beyond category-level change detection toward object-level maps that preserve instance-level structure over time.

Recent object-level semantic mapping methods~\cite{gu2024conceptgraphs,Wu2024HOVSG} represent scenes as collections of RGB-derived object features using foundation models~\cite{Radford2021CLIP,Kirillov2023SAM,ren2024grounded}. This provides open-set, class-agnostic scene representations, which are well suited to long-term environments when the nature of future change is not known in advance. 
However, maintaining longitudinal object-level maps requires reliably associating instances across sessions, which remains the core difficulty.
In real-world inspection scenarios, objects are often occluded (fully or partially), and imperfect segmentation may over-segment a single object or miss it entirely, thereby degrading both object association and change detection.

Existing systems that model object persistence by associating appearance embeddings, spatial proximity, or geometric overlap~\cite{Qian2022POCD,Qian2023POVSLAM,Schmid2024Khronos,Ben2026wheremyglass,saavedra2026predictive} can therefore fail when observations are partial or when visually similar instances co-occur.
Methods that explicitly match pre-segmented instances~\cite{Zhu2024LivingScenes,liu2025sg} require clean, well-reconstructed objects as input, which rarely holds in scene-scale settings with sparse observations and partial views. 
As a result, both types of methods become unreliable in the diverse semi-static environments where long-term inspection matters most.

\begin{figure}
	\centering
	\includegraphics[width=\columnwidth]{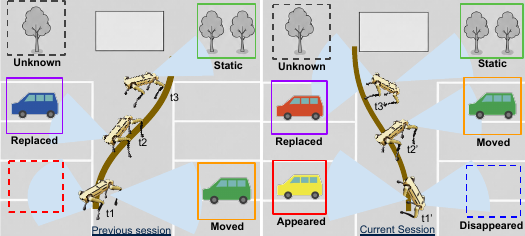}
        \caption{\small{A typical multi-session mapping scenario: a robot revisits the same place (while not precisely repeating the same trajectory). Change detection is performed incrementally within local spatial regions, so the robot first identifies changed objects locally. It accumulates evidence over time to associate changed objects globally. In this example, the green car is initially flagged as having disappeared in current session, but is later re-associated in a new location once enough evidence has accumulated. }}
	\label{fig:method_multi_session_environments}
	\vspace{-1em}
\end{figure}

We present OASIS-map, a multi-session mapping system that maintains a consistent object-level map across revisits under occlusion, partial views, and imperfect segmentation. 
Our key insight is to match image pairs at the patch level rather than comparing whole-object descriptors, enabling both change detection and object association to be driven by fine-grained correspondence evidence. 
Our system first computes dense patch-level correspondences between image pairs. These correspondences are particularly effective when object-level matching is ambiguous due to partial co-visibility, visually similar instances, or viewpoint changes, as they provide confidence over both matched and unmatched regions. This allows changed regions to be identified from unmatched or low-confidence correspondences, while object-level associations are derived from aggregated patch-level evidence.
As illustrated in Fig.~\ref{fig:method_multi_session_environments}, OASIS-Map performs incremental change detection and object association by accumulating dense correspondence evidence over repeated observations. Change labels are confirmed only when sufficient co-visible coverage has been accumulated, rather than being inferred from sparse early views.

The main contributions of our work are as follows:
\begin{enumerate}
    \item \textbf{A unified spatio-temporal object-level mapping system} that fuses object instances, semantic information, and 3D geometry into a consistent map capturing environmental changes across multiple sessions.
    
    \item \textbf{A dense semantic correspondence module} that uses patch-level matches from DINOv3 features~\cite{simeoni2025dinov3} to detect changed image regions and determine object association under partial observations, occlusion, and imperfect segmentation.

    \item \textbf{Evaluation on three real-world sequences} covering indoor object rearrangement (3RScan), geometrically ambiguous object replacement (Car Park), and large-scale outdoor change over days to months (Market).
\end{enumerate}

\section{RELATED WORK}
\label{sec:related_works}

\subsection{Long-term and Multi-session 3D Mapping}

Traditional SLAM systems typically assume a static environment, as is the case in visual~\cite{MurArtal2017ORBSLAM2}, visual-inertial~\cite{qin2018vins} and LiDAR~\cite{wisth2022vilens} SLAM.
Extending single-session SLAM to long-term operation introduces the problem of associating repeat traversals to one another despite changes such as illumination, weather, season, viewpoint, and scene dynamics.
Multi-session mapping methods~\cite{schneider2018maplab,kim2022ltmapper} address this by using place recognition to help register visits into a shared coordinate frame and then consolidating the individual maps into a unified map.

Assuming poses are known in a shared frame, change detection operates over the co-visible regions between sessions — most directly by comparing observations geometrically in 2D or 3D.
In 2D, image-based methods compare observed images against a prebuilt 3D model to identify geometric differences~\cite{taneja2011imageCD,Palazzolo2018fastCD}.  
In 3D, change detection methods typically build a volumetric or geometric representation of the scene and compare it across sessions at the voxel, point, or surface level. Using depth cameras, TSDF-based methods construct and compare signed distance fields to detect changes~\cite{Fehr2017TSDF}. Similarly, LiDAR-based approaches typically identify changes by computing voxel-level differences through occupancy grids, distance metrics, or ephemerality scores~\cite{Rowell2024LiSTA,kim2022ltmapper,gil2025elite}.
However, these methods remain sensitive to localization errors, sensor noise, and map resolution. As a result, detected changes are often noisy, as apparent differences may be caused by misalignment or partial overlap rather than genuine change. More fundamentally, they operate on geometry alone, so they can indicate that a region has changed, but not what has changed.
\subsection{Object-level Semantic Mapping and Scene Graph}
Recent works such as Panoptic Multi-TSDF~\cite{Schmid2022PanopticMultiTSDF} and Hydra~\cite{Hughes2022Hydra} move beyond geometry-only representations by integrating dense volumetric maps with semantic information while preserving the underlying geometry. This combination of geometry and semantics provides a meaningful basis for scene understanding, since semantically distinct changes may be difficult to separate using geometry alone.
On the other hand, object-centric mapping methods such as 3D scene graphs~\cite{gu2024conceptgraphs,Wu2024HOVSG} map the environment as a collection of objects and their relations.
These 3D scene-graph methods often rely on foundation models such as CLIP~\cite{Radford2021CLIP}, DINO~\cite{simeoni2025dinov3}, and SAM~\cite{Kirillov2023SAM} to provide open-set, class-agnostic perception. This makes them well suited to diverse environments where future object categories and changes are not known in advance. A further advantage is that scene graphs are lightweight and easy to update, which makes them a practical representation for capturing object interactions, relations, and state changes~\cite{looper22vsg,behrens2025lostfound,saavedra2026predictive}.
\subsection{Object-level Change Detection and Association}
Some older works focusing on change and persistency estimation include ~\cite{krajnik2017fremen,rosen2016towards}. These methods accumulate repeated and periodic observations and assign a persistence belief to low-level scene elements such as voxels or point features over time. More recent object-level extensions~\cite{Qian2022POCD,Qian2023POVSLAM,Schmid2024Khronos,Ben2026wheremyglass,saavedra2026predictive} lift this idea to objects and track whether each object persists, appears, or disappears across revisits.
However, these methods still depend on reliable associations across sessions, which is difficult when there is a large temporal gap as objects will have moved, been replaced, or been only partially observed.
This exposes a limitation of association in SLAM: most data association is still \textit{sparse} and \textit{implicit}, relying on object embeddings, keypoints, or low-level features that are useful for tracking but are not designed to detect change. In mapping settings, such cues are often insufficient to determine whether two observations should be associated or whether a true correspondence is missing. Stronger change reasoning therefore requires denser signals that integrate geometry, semantics, and object structure.
To address this more explicitly, object-centric matching methods~\cite{Zhu2024LivingScenes,liu2025sg} learn object-shape embeddings for registration and association, while 4D segmentation methods~\cite{steiner2026rescene4d} track instances across a scene over time.
However, these methods assume clean, pre-segmented 3D instances, which suits \textit{object-centric}, room-scale scenes but breaks down in large, \textit{scene-centric} settings.
Real-world applications are more commonly scene-centric, including indoor rescans~\cite{Wald2020RIO10}, inventory monitoring in warehouses~\cite{Park2021ChangeSim}, and construction site inspection~\cite{Sun2023NSS}. These scenes are typically cluttered and undergo structural change, making consistent association across revisits difficult.

\section{METHOD}
\label{sec:Method}

\begin{figure}[h]
 \centering
 \includegraphics[width=\columnwidth]{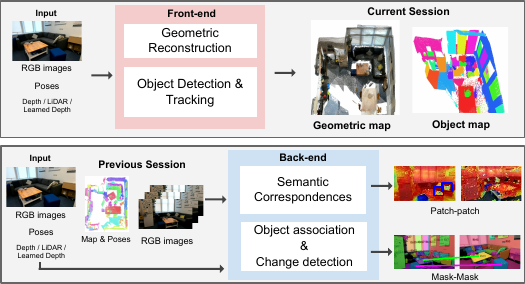}
 \caption{System overview of our multi-session semantic mapping pipeline. \textbf{Front-end} (red): given RGB images, poses, and depth from raw sensors (LiDAR, depth cameras) or learned estimators~\cite{depthanything3}, the front-end builds a geometric map and an object-level semantic map per session using vision foundation models~\cite{ravi2024sam2,simeoni2025dinov3}. \textbf{Back-end} (blue): given the current session object map and the stored previous session map, the back-end establishes dense semantic correspondences between bi-temporal images and performs object-level association to assign change labels $\ell \in \{\texttt{static},\ \texttt{appeared},\ \texttt{disappeared},\ \texttt{moved},\ \texttt{unknown} \}$.}
 \label{fig:pipeline}
 	\vspace{-1em}

\end{figure}

\subsection{System Overview}
An overview of our system is illustrated in Fig.~\ref{fig:pipeline}. The system maintains a spatio-temporal object-level semantic map across multiple mapping sessions. The pipeline consists of two core modules: a front-end object mapping module and a back-end correspondence and association module.

\noindent\textbf{Inputs.} The system takes two mapping sessions as input: a previous, completed session $\mathcal{S}_0$ and a current, ongoing session $\mathcal{S}_1$. Each session provides a stream of keyframes with associated 3D poses and depth. For a keyframe $t \in {1, \dots, N_0}$ in $\mathcal{S}_0$ and $t \in {1, \dots, N_1}$ in $\mathcal{S}_1$, let $\mathbf{I}_t^{(0)}, \mathbf{I}_t^{(1)} \in \mathbb{R}^{H \times W \times 3}$ denote the RGB images, $\mathbf{D}_t^{(0)}, \mathbf{D}_t^{(1)} \in \mathbb{R}^{H \times W}$ the depth maps, and $\mathbf{T}_t^{(0)}, \mathbf{T}_t^{(1)} \in \mathrm{SE}(3)$ the camera poses. Depth may be obtained from a depth sensor, a LiDAR projection, or a learned depth estimator~\cite{depthanything3}.
All poses
across both sessions are expressed in a common, metric world coordinate frame,
obtained separately using a multi-session LiDAR SLAM system~\cite{wisth2022vilens} that aligns the sessions into a shared frame.

\noindent\textbf{Front-end: Object-level Semantic Mapping.} The front-end runs
online and incrementally builds a per-session 3D object map
$\mathcal{O}_s = \{o_s^1, \dots, o_s^{K_s}\}$, where $o_s^k$ is the $k$-th object
instance and $K_s$ the number of instances in session $s$. For each keyframe, the
object detection and tracking module extracts a set of instance masks
$\mathcal{M}_t^{(s)}$ and a dense semantic feature map
$\mathbf{F}_t^{(s)} \in \mathbb{R}^{H_f \times W_f \times d}$ on a low-resolution patch grid, with $d$ the feature dimension; pooling
$\mathbf{F}_t^{(s)}$ over each mask gives a per-object feature vector
$\mathbf{f} \in \mathbb{R}^{d}$. At the end of a session, the front-end outputs a geometric map $\mathcal{V}_s$ together with the object-level semantic map $\mathcal{O}_s$.

\noindent\textbf{Back-end: Correspondence and Association.} Given the output map from a
previous session $( \mathcal{V}_0, \mathcal{O}_0, \mathcal{M}_t^{(0)}, \mathbf{F}_t^{(0)},
\mathbf{T}_t^{(0)})$ and the current observations, the back-end computes
dense patch correspondences $\mathcal{C}^*$ for each co-visible image pair
(Sec.~\ref{sec:dense_semantic_correspondences}), then aggregates them within object
masks into a score matrix $\mathbf{S}$, and extracts a one-to-one association
$\mathcal{A} \subseteq \mathcal{O}_1 \times \mathcal{O}_0$. Once evidence has accumulated within a submap, each object is assigned a change
label $\ell(o)$: matched pairs are labelled \texttt{static} or \texttt{moved},
confidently unmatched objects are \texttt{appeared} or \texttt{disappeared}, while the
rest stay \texttt{unknown} until further observations resolve them.

\subsection{Front-end: Object-level semantic mapping}
\subsubsection{Geometric Reconstruction}

For each session, we reconstruct a local geometric submap every $N_{\text{sub}}{=}10$ keyframes. Depth $\mathbf{D}_t$ (e.g., from a depth camera or a learned estimator such as Depth Anything~\cite{depthanything3}) is integrated into a TSDF voxel map, suppressing sensor noise and scale inconsistencies, while LiDAR point clouds are accumulated directly under $\mathbf{T}_
t$, as their measurements are already metrically accurate. Keyframes are selected adaptively: a new keyframe is added to the buffer when its pose differs from the previous one by more than a translation threshold $\delta_x$ or a rotation threshold $\delta_\theta$, set according to the depth modality — $(\delta_x{=}0.5\,\text{m}, \delta_\theta{=}20\degree)$ for raw depth sensors and $(\delta_x{=}1.0\,\text{m}, \delta_\theta{=}30\degree)$ for learned estimators, where the wider baseline compensates for scale-inconsistent per-frame predictions by providing greater multi-view overlap. The pipeline outputs a combined full geometric map $\mathcal{V}_s$ at the end of the session, expressed in the shared world frame.

\subsubsection{Object detection and tracking}
To detect and track objects, we apply SAM2~\cite{ravi2024sam2} to each RGB
keyframe to obtain instance masks, and then the DINOv3~\cite{simeoni2025dinov3}
encoder $g$ to extract a dense feature map
$\mathbf{F}_t = g(\mathbf{I}_t) \in \mathbb{R}^{H \times W \times d}$. The
candidate masks detected across the session's keyframes are lifted into 3D using
the depth $\mathbf{D}_t$ and pose $\mathbf{T}_t$. Following
ConceptGraphs~\cite{gu2024conceptgraphs}, overlapping 3D segments are merged
across keyframes when both their geometric overlap and semantic similarity
exceed fixed thresholds. When 3D merging is insufficient (for instance, in low-FoV scenes where segments are only partially observed), we additionally track objects in 2D by projecting the 3D object points onto the image plane and measuring their overlap with the detected masks.

Instead of averaging DINOv3 features across views, which discards view-specific discriminative information, we store the per-keyframe dense feature maps $\mathbf{F}_t$ together with per-object feature tokens cropped from each keyframe. These are used later for semantic correspondence matching (Sec.~\ref{sec:dense_semantic_correspondences}).
The front-end output for session $s$ consists of the geometric submaps, the
object map $\mathcal{O}_s = \{o_s^1, \dots, o_s^{K_s}\}$, the per-keyframe dense
feature maps $\{\mathbf{F}_t\}_{t=1}^{N_s}$, and the poses
$\{\mathbf{T}_t\}_{t=1}^{N_s}$, all passed to the back-end.

\subsection{Back-end: Correspondence and Association}

\subsubsection{Dense semantic correspondences}
\label{sec:dense_semantic_correspondences}

For a co-visible pair of feature maps from the two sessions, $\mathbf{F}_0$
from $\mathcal{S}_0$ and $\mathbf{F}_1$ from $\mathcal{S}_1$, reused from the
front-end and flattened to $\mathbf{F}_0, \mathbf{F}_1 \in
\mathbb{R}^{H_f W_f \times d}$. 
We form the correlation matrix between patch $p$ of $\mathbf{F}_0$ and patch $q$ of
$\mathbf{F}_1$
\begin{equation}
\mathbf{C} = \mathbf{F}_0 \mathbf{F}_1^{\top}
\in \mathbb{R}^{H_f W_f \times H_f W_f}, \qquad
\mathbf{C}_{pq} = \big\langle \mathbf{F}_0(p),\, \mathbf{F}_1(q) \big\rangle,
\end{equation}
From the same matrix we take, for every patch in $\mathcal{S}_0$ and $\mathcal{S}_1$, its best match in
each direction --- the forward match over each row and the backward match over
each column:
\begin{equation}
\hat{q}(p) = \argmax_{q} \mathbf{C}_{pq}, \qquad
\hat{p}(q) = \argmax_{p} \mathbf{C}_{pq}.
\end{equation}
Each retained match carries a confidence score given by its correlation entry,
\begin{equation}
s(p,q) = \mathbf{C}_{pq}.
\label{eq:conf}
\end{equation}
We keep the best match in both directions --- forward from $\mathcal{S}_0$ to
$\mathcal{S}_1$ and backward from $\mathcal{S}_1$ to $\mathcal{S}_0$ --- giving the
correspondence set
\begin{equation}
\mathcal{C}^* =
\underbrace{\{(p,\hat{q}(p))\}_{p}}_{\mathcal{S}_0 \to \mathcal{S}_1}
\;\cup\;
\underbrace{\{(\hat{p}(q),q)\}_{q}}_{\mathcal{S}_1 \to \mathcal{S}_0},
\label{eq:corr_set}
\end{equation}
where $p$ ranges over patches of $\mathbf{F}_0$ and $q$ over patches of
$\mathbf{F}_1$.

\begin{figure}[t]
	\centering
	\includegraphics[width=\columnwidth]{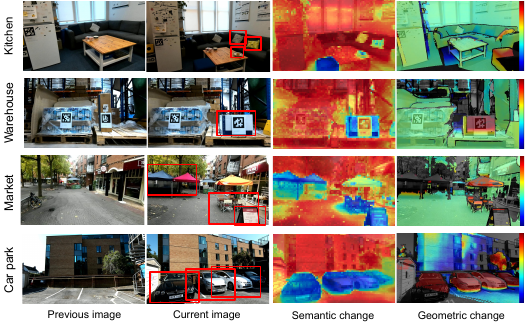}
        \caption{\small{Change detection using semantic and geometric cues. Given a pair of bi-temporal images captured from nearby viewpoints, our change detection module produces both semantic and geometric change heatmaps. \textbf{Semantic change} (third column) is derived from DINO patch correspondences, where red denotes persistent regions with high feature similarity and blue highlights changed regions characterised by low similarity or unmatched patches. \textbf{Geometric change} (fourth column) visualises signed distance differences, where red ($+$) indicates newly appearing foreground regions and blue ($-$) indicates regions that are no longer present.}}
	\label{fig:method_change_2d}
	\vspace{-1em}
\end{figure}

\subsubsection{Object-level association}

Given the dense correspondences $\mathcal{C}^*$, a previous-session object
$o_0^l \in \mathcal{O}_0$ and a current-session object $o_1^k \in \mathcal{O}_1$, we
associate them using their image masks $M_0^l$ and $M_1^k$.

\noindent\textbf{\emph{Patch-to-mask aggregation.}}
For a previous mask $M_0^l$ and a current mask $M_1^k$, we gather the matches leaving $M_0^l$, and the subset that lands in $M_1^k$:
\begin{equation}
\begin{aligned}
\mathcal{C}_{l}  &= \underbrace{\{(p,q)\in\mathcal{C}^* \mid p\in M_0^l\}}_{M_0^l\,\to\,\cdot}, \\[2pt]
\mathcal{C}_{lk} &= \underbrace{\{(p,q)\in\mathcal{C}^* \mid p\in M_0^l,\ q\in M_1^k\}}_{M_0^l\,\to\,M_1^k}.
\end{aligned}
\end{equation}

\noindent\textbf{\emph{Mask-to-mask matching.}}
We score each object pair $(l,k)$ by two quantities over its linking matches
$\mathcal{C}_{lk}$: the mean confidence $\bar{s}_{lk}$, and the match proportion
$\sigma_{lk}$ --- the fraction of $M_0^l$'s matches that reach $M_1^k$:
\begin{equation}
\bar{s}_{lk} =
\frac{1}{|\mathcal{C}_{lk}|}\!\!\sum_{(p,q)\in\mathcal{C}_{lk}}\!\! s(p,q),
\qquad
\sigma_{lk} = \frac{|\mathcal{C}_{lk}|}{|\mathcal{C}_l|}.
\label{eq:mask_scores}
\end{equation}
We then analyse the distribution of correspondence scores within each mask to distinguish 
reliable associations from ambiguous ones. Four cases arise: 
\begin{enumerate}
    \item \textit{high score, high match proportion} --- patches match confidently and 
densely, indicating a true positive association; 
    \item \textit{high score, low match proportion} --- only a sparse subset of patches 
match confidently, typically occurring when objects are partially visible or occluded 
across sessions; 
    \item \textit{low score, high match proportion} --- many patches find nearest-neighbour 
matches but with low confidence, indicating false positive matches between semantically 
similar objects like between different cars or boxes; 
    \item \textit{low score, low match proportion} --- neither confident nor dense matches 
exist, indicating the object has no reliable counterpart in the other session.
\end{enumerate}
As shown in Fig.~\ref{fig:method_change_2d}, case~3 is what
lets us distinguish visually similar but distinct instances: two different boxes in
a warehouse yield a high match proportion ($\sigma_{lk}$) yet low confidence
($\bar{s}_{lk}$). Because we score correspondences at the patch level rather than
collapsing each object into a single global descriptor, this low-confidence signal
is preserved and the two instances can be told apart.
We therefore accept a confident association only in case~1, thresholding $\bar{s}_{lk}$
and $\sigma_{lk}$ jointly; case~2 (partial visibility) is deferred and resolved later as
more co-visible evidence accumulates; while cases~3 and~4 are rejected.

Association is performed locally within a submap of $N_{\text{sub}}$ keyframes and
then globally. For the object sets $\mathcal{O}_0 = \{o_0^1, \dots, o_0^n\}$ and
$\mathcal{O}_1 = \{o_1^1, \dots, o_1^m\}$ observed within the submap, we construct
the pairwise score matrix
\begin{equation}
\mathbf{S} \in \mathbb{R}^{n \times m}, \qquad \mathbf{S}_{lk} = \sigma_{lk}.
\end{equation}
We compute a one-to-one assignment $\mathcal{H}$ between $\mathcal{O}_0$ and
$\mathcal{O}_1$ by solving the Hungarian algorithm on $\mathbf{S}$, and keep a matched pair only when enough of an object matches
($\sigma_{lk} \geq \tau_\sigma$) and the matches are confident
($\bar{s}_{lk} \geq \tau_s$), which together retain only case~1:
\begin{equation}
\mathcal{A} = \bigl\{ (o_0^l, o_1^k) \in \mathcal{H} \mid
\sigma_{lk} \geq \tau_\sigma,\ \bar{s}_{lk} \geq \tau_s \bigr\}.
\end{equation}

\subsubsection{Change label assignment}
Once evidence has accumulated within a submap, each object is assigned a change label directly from the association outcome:
\begin{equation}
\ell(o) =
\begin{cases}
\texttt{static}      & (o_0^l, o_1^k) \in \mathcal{A},\ \|\mathbf{c}_1^k - \mathbf{c}_0^l\| \leq \tau_d \\
\texttt{moved}       & (o_0^l, o_1^k) \in \mathcal{A},\ \|\mathbf{c}_1^k - \mathbf{c}_0^l\| > \tau_d \\
\texttt{appeared}    & o_1^k \text{ unmatched},\ \nu(o_1^k) \geq \nu_{\min} \\
\texttt{disappeared} & o_0^l \text{ unmatched},\ \nu(o_0^l) \geq \nu_{\min} \\
\texttt{unknown}     & \text{otherwise}
\end{cases}
\label{eq:change_labels}
\end{equation}
where $\mathbf{c}_s^k \in \mathbb{R}^3$ is the centroid of object $o_s^k$ and
$\tau_d$ a displacement threshold. Here $\nu(o) \in [0,1]$ is the 3D coverage of an
object --- the fraction of its volume that lies within the region observed by the
other session --- and we declare \texttt{appeared} or \texttt{disappeared} only
when more than half the volume is co-visible ($\nu_{\min} = 0.5$); otherwise the
object is left \texttt{unknown}.

\begin{figure*}[t]
	\centering
	\includegraphics[width=\textwidth]{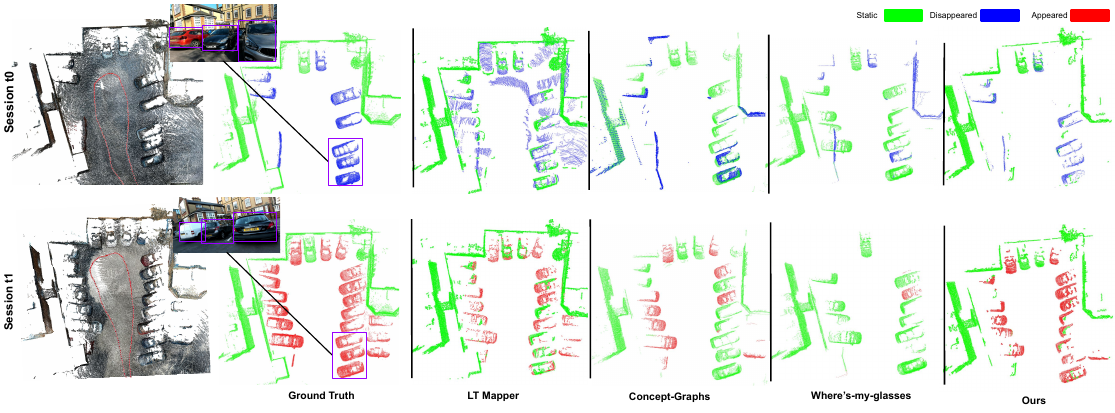}
	\caption{\small{Qualitative results on 3D change detection. Ground-truth change labels are shown as coloured points: static (green), e.g.\ the building and parked cars; disappeared (blue) and appeared (red), mostly cars. Purple boxes highlight replaced cases,
    where a car appeared and disappeared in the same spatial location. }}
	\label{fig:eval_carpark}
\end{figure*}

\section{EXPERIMENTAL RESULTS}
\label{sec:experiments}

\textbf{Datasets.} We evaluate our approach on real-world indoor and outdoor sequences. \textit{3RScan}~\cite{Wald2020RIO10} is a public indoor dataset in which objects are moved and rearranged within a single room captured with noisy RGB-D data and partial observations from a low-FoV camera. The custom \textit{Car Park} sequence presents geometrically ambiguous cases in which the same spatial region is occupied by different objects across mapping sessions; such object replacements are challenging for geometry-based change detection methods. Finally, the \textit{Market} sequence serves as a qualitative real-world inspection demonstration, capturing a larger-scale outdoor scene where scenes are evolving as shown in Fig.~\ref{fig:header}. We report quantitative and qualitative results on \textit{Car Park} and \textit{3RScan}. For the custom car park dataset, we manually annotate ground-truth object-level changes directly in the 3D point clouds, assigning each instance a class and a change label (\texttt{static}, \texttt{appeared}, \texttt{disappeared}, \texttt{moved}) across sessions. For the 3RScan dataset, we re-annotate some small-object changes that were omitted from the original 3RScan labels, because the original 3D segmentations are sometimes too coarse to capture fine-grained changes that are visible in RGB images.

\textbf{Metrics.} Using the annotated ground truth, we conduct two sets of experiments. (1) 3D change detection: we compare predicted point-wise change labels against ground-truth annotations and report Precision, Recall, and F1-score for each change category $\ell \in \{\texttt{static},\ \texttt{appeared},\ \texttt{disappeared},\ \texttt{moved}\}$, with results also visualized qualitatively as 3D change maps. (2) object detection and association: we evaluate object detection at the end of each session and the consistency of object identity across sessions, which is crucial for object-level spatio-temporal mapping. Other change types such as texture, deformation, or internal state changes (e.g., a door opening)  are excluded from evaluation.

\textbf{Baselines.} We compare against three baseline systems. \textit{LT-Mapper}~\cite{kim2022ltmapper} is a geometry-based method that detects changes via raw point-to-point distances across sessions. 
\textit{ConceptGraphs}~\cite{gu2024conceptgraphs} is an open-vocabulary object-level mapping method which assumes a static scene. We therefore modified it to support multi-session comparison by matching and associating objects offline at the end of each session. \textit{Where's-my-glasses}~\cite{Ben2026wheremyglass} employs an open-set object-level probabilistic framework to estimate each object's state from observations across sessions.

\begin{table*}[t]
\centering
\caption{\small{3D change detection on Car Park and 3RScan. The \textit{Obj.} column indicates whether a method represents objects, and \textit{Assoc.} whether it maintains object identity across sessions. 
Car Park features \textit{replaced} objects; 3RScan features \textit{moved} objects. N.A. marks change types that do not occur in a dataset; – marks a change type a method cannot produce (LT-Mapper has no object information, thus moved objects cannot be detected) }}
\label{tab:eval_3d_change}
\footnotesize
\setlength{\tabcolsep}{4pt}
\renewcommand{\arraystretch}{1.15}
\begin{tabular}{ll cc | ccc | ccc | ccc | ccc}
\hline
& & & & \multicolumn{3}{c|}{Static}
& \multicolumn{3}{c|}{Appear / Disappear}
& \multicolumn{3}{c|}{Replaced}
& \multicolumn{3}{c}{Moved} \\
Dataset & Method & \textit{Obj}. & \textit{Assoc.}
& P$\uparrow$ & R$\uparrow$ & F1$\uparrow$
& P$\uparrow$ & R$\uparrow$ & F1$\uparrow$
& P$\uparrow$ & R$\uparrow$ & F1$\uparrow$
& P$\uparrow$ & R$\uparrow$ & F1$\uparrow$ \\
\hline
\multirow{4}{*}{Car Park}
 & LT-Mapper & \ding{55} & \ding{55}
& 0.563 & 0.440 & 0.438
& \textbf{0.828} & \textbf{0.687} & \textbf{0.748}
& 0.817 & 0.229 & 0.324
& \multicolumn{3}{c|}{N.A.} \\
 & ConceptGraphs & \ding{51} & \ding{55}
& \textbf{0.815} & \underline{0.530} & \underline{0.619}
& 0.560 & 0.243 & 0.281
& \underline{0.876} & \underline{0.601} & \underline{0.686}
& \multicolumn{3}{c|}{N.A.} \\
 & Where's-my-glasses & \ding{51} & \ding{51}
& 0.445 & 0.362 & 0.383
& 0.383 & 0.141 & 0.170
& 0.418 & 0.506 & 0.446
& \multicolumn{3}{c|}{N.A.} \\
 & OASIS-Map (Ours) & \ding{51} & \ding{51}
& \underline{0.699} & \textbf{0.788} & \textbf{0.736}
& \underline{0.682} & \underline{0.616} & \underline{0.592}
& \textbf{0.924} & \textbf{0.732} & \textbf{0.783}
& \multicolumn{3}{c|}{N.A.} \\
\hline
\multirow{4}{*}{3RScan}
 & LT-Mapper & \ding{55} & \ding{55}
& \underline{0.714} & \textbf{0.904} & \textbf{0.798}
& \textbf{0.204} & \textbf{0.686} & \textbf{0.314}
& \multicolumn{3}{c|}{N.A.}
& \multicolumn{3}{c}{--} \\ %
 & ConceptGraphs & \ding{51} & \ding{55}
& 0.686 & 0.476 & 0.562
& 0.177 & \underline{0.362} & 0.238
& \multicolumn{3}{c|}{N.A.}
& \underline{0.359} & \textbf{0.272} & \underline{0.309} \\
& Where's-my-glasses & \ding{51} & \ding{51}
& 0.427 & 0.301 & 0.353
& 0.142 & 0.348 & 0.201
& \multicolumn{3}{c|}{N.A.}
& 0.345 & 0.116 & 0.173 \\
 & OASIS-Map (Ours) & \ding{51} & \ding{51}
& \textbf{0.761} & \underline{0.587} & \underline{0.663}
& \underline{0.195} & 0.314 & \underline{0.241}
& \multicolumn{3}{c|}{N.A.}
& \textbf{0.699} & \underline{0.236} & \textbf{0.353} \\
\hline
\end{tabular}
\end{table*}

\subsection{Evaluation of 3D Change Detection}
Fig.~\ref{fig:eval_carpark} shows change detection results on the Car Park sequence. In Car Park, the same spatial region is frequently occupied by different cars across sessions: a previous-session instance is labelled \textit{disappeared} while a new, co-located instance is labelled \textit{appeared} — together constituting a \textit{replaced} event.
As indicated by the \textit{Obj.} and \textit{Assoc.} columns of Tab.~\ref{tab:eval_3d_change}, LT-Mapper~\cite{kim2022ltmapper} uses neither objects nor their identity across sessions. ConceptGraphs~\cite{gu2024conceptgraphs} is object-level but associates objects only offline. Only Where's-my-glasses shares both capabilities with our method and is therefore the only directly comparable baseline.
LT-Mapper mixes static, appeared, and disappeared labels within the \textit{replaced} region rather than capturing the replacement as one event. ConceptGraphs and Where's-my-glasses both mis-associate the new instance with the old one and label it \textit{static}. Our method instead detects the entire car as a single changed object, correctly separated from the old instance.
Tab.~\ref{tab:eval_3d_change} illustrates this quantitatively. Even though LT-Mapper achieves a high F1-score on appear and disappear (0.748), it cannot handle replaced objects (0.324).
ConceptGraphs associates objects offline and one-to-one at the end of each session, so any object it fails to match is directly considered as appeared or disappeared; under over-segmentation or detection failure, a persistent object that is simply not visible in one session is left unmatched and wrongly reported as disappeared, yielding a low appear/disappear F1 (0.281). 
Where's-my-glasses associates objects online, but relies solely on CLIP embeddings, which is ambiguous when distinguishing one car from another: replaced cars are mistaken for the same instance, while static objects are incorrectly split into a disappearance and an appearance, giving low F1 across static (0.383), appear/disappear (0.170), and replaced (0.446).
Our method also associates objects incrementally across sessions, but using dense, patch-level semantic correspondences rather than a single per-object embedding, remaining robust under partial views and inconsistent segmentation. This lets our method correctly recognise a replaced car as a different instance, achieving the best F1 across all baselines on both replaced (0.783) and static (0.736) objects. 
Since LT-Mapper does not support object-level representation, we exclude it from the remaining experiments and evaluate only the rest of methods in the following object detection and association analysis.

\begin{figure*}[t]
    \centering
    \includegraphics[width=\textwidth]{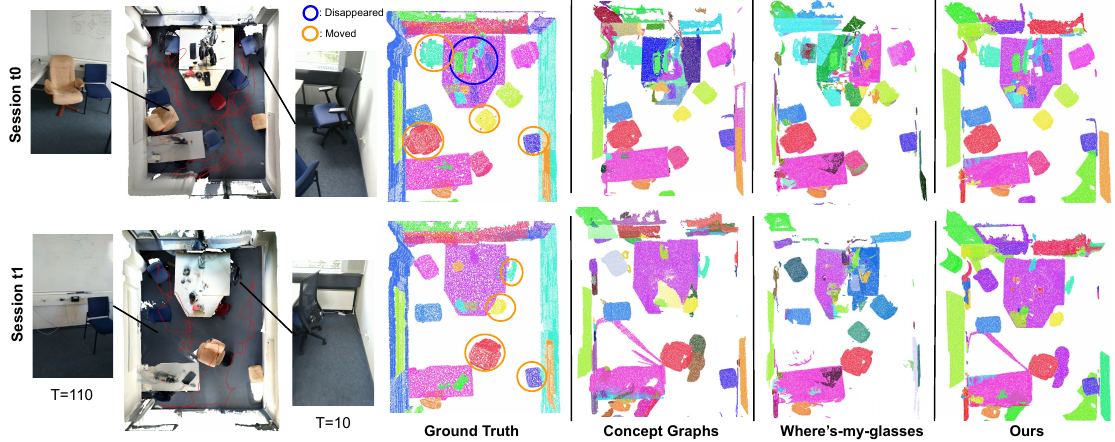}
    \caption{\small{Qualitative results on the 3RScan sequence. In the ground truth (GT), objects are coloured by identity. Persistent
objects (static or moved) keep the same colour across sessions, whereas appeared
and disappeared objects receive unique colours.
Between sessions, the chairs are rearranged and small objects such as monitors and keyboards are removed from the tables. Our method reproduces the ground-truth identities: persistent objects retain a consistent colour across sessions and disappeared objects are correctly flagged, yielding a spatio-temporally consistent map. In contrast, CG and WMG mis-associate objects, visible as inconsistent colours across
sessions}}
    \label{fig:3rscan_qualitative}
    \vspace{-1em}
\end{figure*}

\subsection{Object detection and association}
Table~\ref{tab:detection_result} reports object detection results at the end of each session, measuring the proportion of ground-truth objects discovered (recall) and the proportion of detected objects that are correct (precision) penalising false positives introduced by over-segmentation.
Since the same object detection and segmentation models are used for all methods, the most meaningful comparison is whether the mapping systems can maintain a higher proportion of ground-truth objects (recall) and can preserve them across sessions. 
We argue that changes in the environment can occur at different object granularities; for example, in 3RScan, the main changes involve chairs and small objects on tables, which are typically hard to detect in the initial session. Overall, our method discovers most of the objects in $S_0$ (29 out of 32), achieving the highest recall (0.906) among all methods. It then selectively maintains the persistent objects in $S_1$ (13 out of 21) with recall (0.619).
Table~\ref{tab:association_result} reports object association performance, measuring how object identity is preserved between $S_0$ and $S_1$.
ConceptGraphs~\cite{gu2024conceptgraphs} processes the two sessions independently and associates objects offline, so any unmatched object is simply labelled disappeared in $S_0$ and appeared in $S_1$ --- e.g., small table-top objects missed in $S_0$ vanish from the map in $S_1$ (Fig.~\ref{fig:3rscan_qualitative}). However, the matching is not robust under inconsistent segmentation: a table split into several parts in one session but detected as one in the other breaks the association. It also cannot distinguish similar moved objects such as chairs, giving low recall on static (0.154) and moved (0.154) objects (Tab.~\ref{tab:association_result}).
Where's-my-glasses~\cite{Ben2026wheremyglass} detects changes incrementally while mapping the scene. However, its object association relies on per-object CLIP embeddings. It still struggles to distinguish visually similar objects such as chairs, resulting in low recall (0.167). In addition, missed or partial detections are interpreted as object disappearance, which causes static objects to be incorrectly classified as removed, leading to low recall (0.154).
In contrast, our method first detects changes locally via dense semantic correspondences and then re-associates objects using both geometric cues (e.g., moved chairs) and semantic cues (e.g., small objects on a table), making it robust to detection failures and visual ambiguity. As shown in Tab~\ref{tab:association_result}, this yields higher recall on both \emph{static} (0.385) and \emph{moved} (0.500) objects than other baseline methods.

\begin{table}[t]
\centering
\caption{\small{Object detection evaluation on the 3RScan dataset. 
Each method is evaluated at the end of every session ($S_0$: previous session, $S_1$: current session). 
TP denotes a correct one-to-one match with a ground-truth object, 
FN a ground-truth object that was missed (not detected), 
and FP a predicted object with no corresponding ground-truth (e.g., over-segmentation). }
}
\label{tab:detection_result}
\setlength{\tabcolsep}{3pt}
\renewcommand{\arraystretch}{1.2}
\resizebox{\columnwidth}{!}{%
\begin{tabular}{llcccccc}
\toprule
\textbf{Scan} & \textbf{Methods} & TP $\uparrow$ & FN $\downarrow$ & FP $\downarrow$ & P $\uparrow$ & R $\uparrow$ & F1 $\uparrow$ \\
\midrule
\multirow{4}{*}{$S_0$ (reference, $N_{gt}=32$)}
 & ConceptGraphs     & 21 & 11 & 13 & 0.618 & 0.656 & 0.636 \\
 & Where's-my-glasses  & 22 & 10 & 21 & 0.512 & 0.688 & 0.587 \\
 & OASIS-Map (Ours)              & \textbf{29} & \textbf{3}  & 21 & 0.580 & \textbf{0.906} & \textbf{0.707} \\
\midrule
\multirow{4}{*}{$S_1$ (rescan, $N_{gt}=21$)}
 & ConceptGraphs     & \textbf{13} & 8  & 11 & 0.542 & \textbf{0.619} & 0.578 \\
 & Where's-my-glasses  & 11 & 10 & \textbf{5}  & \textbf{0.688} & 0.524 & \textbf{0.595} \\
 & OASIS-Map (Ours)              & \textbf{13} & 8  & 15 & 0.464 & \textbf{0.619} & 0.531 \\
\bottomrule
\end{tabular}%
}
\end{table}

\begin{table}[t]
\centering
\caption{\small{Object association evaluation on the 3RScan dataset. Three change categories are evaluated: static, moved and removed objects.}}
\label{tab:association_result}
\setlength{\tabcolsep}{2pt}
\renewcommand{\arraystretch}{1.05}
\scriptsize
\begin{tabular}{l|ccc|ccc|ccc}
\toprule
& \multicolumn{3}{c|}{Static ($N_{gt}=13$)} 
& \multicolumn{3}{c|}{Moved ($N_{gt}=12$)} 
& \multicolumn{3}{c}{Removed ($N_{gt}=11$)} \\
\textbf{Method} 
& P$\uparrow$ & R$\uparrow$ & F1$\uparrow$
& P$\uparrow$ & R$\uparrow$ & F1$\uparrow$
& P$\uparrow$ & R$\uparrow$ & F1$\uparrow$ \\
\midrule
ConceptGraphs 
& 1.000 & 0.154 & 0.267
& 1.000 & 0.154 & 0.267
& \textbf{0.909} & 1.000 & \textbf{0.952} \\

Where's-my-glasses 
& 0.500 & 0.154 & 0.235
& 1.000 & 0.167 & 0.286
& 0.364 & 1.000 & 0.533 \\

OASIS-map (Ours) 
& 1.000 & \textbf{0.385} & \textbf{0.556}
& 1.000 & \textbf{0.500} & \textbf{0.667}
& 0.800 & 0.889 & 0.842\\
\bottomrule

\end{tabular}
\end{table}

\subsection{Computation time and Memory}
We evaluate our method on an NVIDIA RTX 4090 (24~GB). \textit{Front-end:} Runs per keyframe with the following breakdown:
DINOv3 feature extraction ($\sim$10\,ms), object detection and 
segmentation ($\sim$100\,ms for 10 objects), and 
TSDF reconstruction ($\sim$50\,ms); every $N_\text{sub}=10$ keyframes it  additionally builds a submap and performs object postprocessing in $\sim$100\,ms. On the Car Park sequence 
(276 keyframes, 94.7\,s), keyframes are sampled at $\sim$3\,fps, 
comfortably within this budget.
\textit{Back-end:} Performs per-keyframe change detection via 
dense correspondence ($\sim$10\,ms) and dense raycasting ($\sim$100\,ms). 
Association is computed per submap (every $N_\text{sub}=10$ keyframes) in $\sim$100\,ms. Features are reused at negligible cost, but object associations must be recomputed for each submap for newly detected objects.
GPU memory usage remained below 10~GB throughout.

\section{CONCLUSIONS}
In this work, we presented OASIS-Map, a multi-session mapping system that maintains a spatio-temporally consistent object-level map across revisits. OASIS-Map jointly performs change detection and object association using dense semantic correspondences, which remain robust to partial views, occlusion, and visual ambiguity. Evaluation on real-world semi-static environments shows that our method maintains consistent object identities across sessions and reliably 
detects appeared, disappeared, and moved objects. In future work, we will extend the system beyond bi-temporal comparison to large-scale, long-term longitudinal mapping.

\bibliographystyle{IEEEtran}
\bibliography{IEEEabrv, references}

\end{document}